\documentclass[journal]{IEEEtran}
\IEEEoverridecommandlockouts

\usepackage{graphics} 
\usepackage{cite}
\usepackage{amsmath}
\usepackage{graphicx}
\usepackage{hyperref}
\usepackage{xcolor}
\usepackage{soul}
\usepackage{algorithm}
\usepackage{algorithmic}
\usepackage{amsmath,amssymb,amsfonts}
\usepackage[normalem]{ulem}
\usepackage{float}
\usepackage{xcolor}
\usepackage{textcomp}
\usepackage[caption=false]{subfig}

\def\BibTeX{{\rm B\kern-.05em{\sc i\kern-.025em b}\kern-.08em
    T\kern-.1667em\lower.7ex\hbox{E}\kern-.125emX}}

\begin{document}

\title{Enhancing a Neurocognitive Shared Visuomotor Model for Object Identification, Localization, and Grasping With Learning From Auxiliary Tasks
{
\thanks{The authors gratefully acknowledge partial support from the German Research Foundation DFG under project CML (TRR 169).}
}}

\IEEEoverridecommandlockouts
\IEEEpubid{\makebox[\columnwidth]{Accepted for publication in IEEE \href{https://cis.ieee.org/publications/t-cognitive-and-developmental-systems}{TCDS}. ~\copyright2020 IEEE\hfill} \hspace{\columnsep}\makebox[\columnwidth]{!}}

\author{
\IEEEauthorblockN
{Matthias Kerzel\IEEEauthorrefmark{1}, Fares Abawi\IEEEauthorrefmark{1}, Manfred Eppe and Stefan Wermter}

\IEEEauthorblockA{\textit{Knowledge Technology, Department of Informatics, University of Hamburg, Germany}\\
    kerzel / abawi / eppe / wermter @informatik.uni-hamburg.de\\ 
    \url{http://www.knowledge-technology.info}\\
    \IEEEauthorrefmark{1}Both authors contributed equally.
    \vspace{-0.5em}
    }
}

\maketitle

\IEEEpubidadjcol

\newcommand{\squeezefigs}{-0.5em}
\setlength{\belowcaptionskip}{-5pt}


\begin{abstract}
We present a follow-up study on our unified visuomotor neural model for the robotic tasks of identifying, localizing, and grasping a target object in a scene with multiple objects. Our Retinanet-based model enables end-to-end training of visuomotor abilities in a biologically inspired developmental approach. In our initial implementation, a neural model was able to grasp selected objects from a planar surface. We embodied the model on the NICO humanoid robot. In this follow-up study, we expand the task and the model to reaching for objects in a three-dimensional space with a novel dataset based on augmented reality and a simulation environment.
We evaluate the influence of training with auxiliary tasks, i.e., if learning of the primary visuomotor task is supported by learning to classify and locate different objects. We show that the proposed visuomotor model can learn to reach for objects in a three-dimensional space. We analyze the results for biologically-plausible biases based on object locations or properties. We show that the primary visuomotor task can be successfully trained simultaneously with one of the two auxiliary tasks. This is enabled by a complex neurocognitive model with shared and task-specific components, similar to models found in biological systems.

\end{abstract}



\section{INTRODUCTION}

We present a follow-up study to the biologically inspired neural model for the robotic task of object identification, localization, and motor action regression, introduced by Kerzel et al.~\cite{self}. We enhance the approach on three fronts: \textbf{1)} Extend the model's visuomotor capabilities from reaching for objects on a planar surface to reaching objects in a three-dimensional space as this is a required ability for many real-world robotic applications; \textbf{2)} Address the influence of imbalances in the training data on possible biases in the model's behavior; \textbf{3)} Evaluate the model's performance on the auxiliary tasks of object localization and identification. We examine the effect of training these auxiliary tasks, along with the main task of reaching for an object, to gain a better understanding of the model's performance and observe possible synergetic effects of learning the three tasks simultaneously.

In a developing human, learning the visuomotor task of reaching for an object goes hand-in-hand with the visual tasks of localizing and identifying the object in their visual field~\cite{mccarty2001infants}. On a neurocognitive level, these tasks share an important component, namely the early visual system. In the human brain, visual stimuli are processed through shared initial steps before branching off into the more specialized dorsal \emph{where} and ventral \emph{what} pathways~\cite{kruger2013deep}. Attention mechanisms modulate this processing of visual information, i.e., pattern, shape, and kinetic information is filtered by spatial or task-driven goal cues in the dorsal pathway~\cite{SIEGEL2008709DorsalVisualPathway}, thus realizing the fusion of top-down semantic information into the visual processing stream. It can be hypothesized that this shared component benefits from learning all related tasks. These assumptions about the human brain can be realized in an artificial neurocognitive shared visuomotor model, in terms of having a shared early visual processing stream that branches out into task-specific networks.

In previous work, we designed a model for reach-for-grasp actions towards a target object in a scene with multiple objects \cite{self}. The model combined the convolutional component of the Resnet~\cite{he2016deep}, a Feature Pyramid Network (FPN) for object classification and localization~\cite{lin2017focal}, neurally encoded linguistic labels, and feed-forward layers for a robotic reach-for-grasp task~\cite{kerzel2017neural}. The model processed an image from the robot's perspective and predicted joint values to reach for a target object that is specified in terms of semantic features, such as type, color, and shape. We showed that a neurocognitively inspired, unified neural model could learn visuomotor abilities for grasping target objects on a table surface. This model was embodied in a developmental robot, through which grasp failures were observed. These failures were interpreted as grasps targeting the wrong object in a scene, contrary to mis-grasping the correct object. The systematic nature of the observed mis-grasps resembles observations of human grasp development by Libertus et al.~\cite{libertus2013size}, who reported that infants show a reaching preference towards objects with a high general visual salience. We hypothesize that both in the biological as well as our artificial system, the observed behavior can be attributed to the selection of the target object during visual processing, not to the ability to reach for the desired object. For a more detailed analysis of our previous work, see \cite{self}; for a general discussion on the biological plausibility of supervised learning and backpropagation, we refer the reader to~\cite{lillicrap2020backpropagation}.

\begin{figure*}[ht]
\centering
\includegraphics[width=0.32\linewidth]{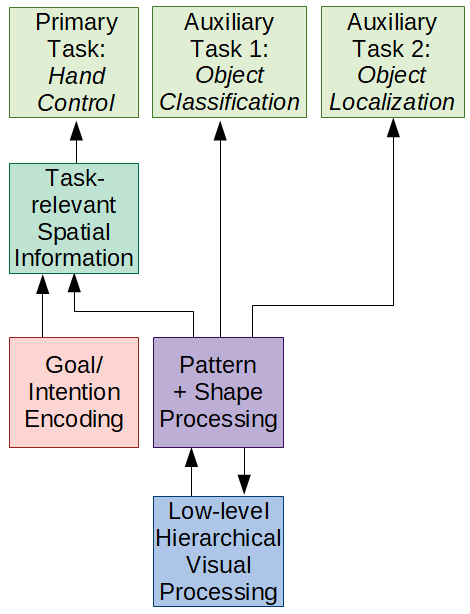}
\hfill
\includegraphics[width=0.32\linewidth]{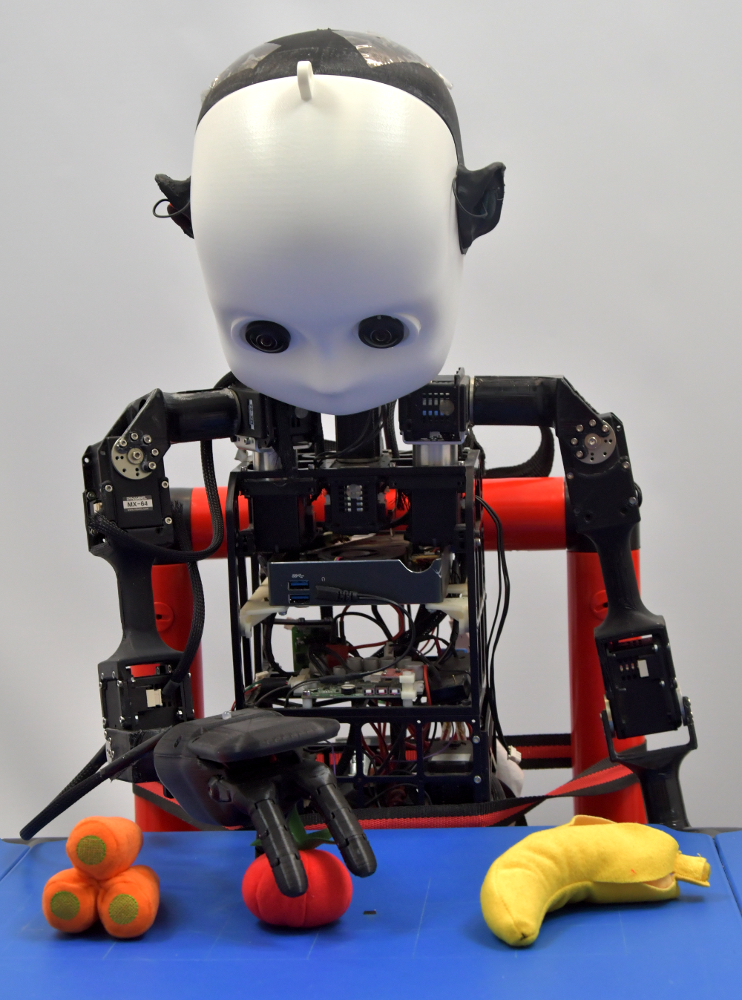}
\hfill
\includegraphics[width=0.3235\linewidth]{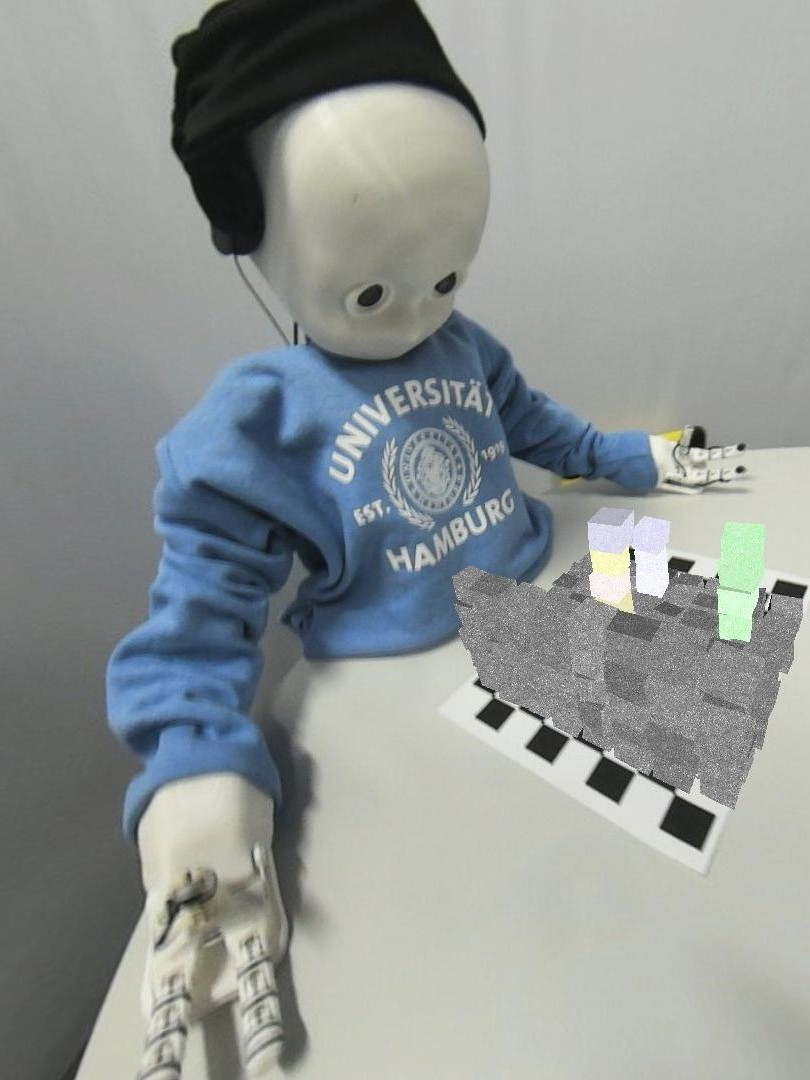}
\caption{Left: Extended End-to-end model for the primary task of grasping a selected object and the two auxiliary tasks of object classification and object localization. Middle: Physical NICO humanoid robot picking a selected object (red tomato) from a desk with two distractor objects (adapted from~\cite{self}). Right: Realization of the experimental augmented reality setup with synthetic blocks overlaid on top of a real image.}
\label{fig:setup}\vspace{\squeezefigs}
\end{figure*}

In this follow-up study, we adapt our model to the novel task of three-dimensional reach-for-grasp and also simultaneous training of the auxiliary tasks of object localization and identification with a combined loss function that incorporates all three tasks, as shown in \autoref{fig:setup} (left).  To this end, a new dataset is generated, based on the Extended Train Robot~\cite{ETR} dataset, the robotic simulation environment MuJoCo\footnote{\url{http://www.mujoco.org/}}, and augmented reality. Due to the high number of required training samples for this more complex task, realizing a real-world dataset is prohibitive. The dataset features 37,500 samples, each consisting of an image with the following annotations: joint configuration to reach for the target object (primary task), position of the selected object in the image (auxiliary task), classification of the image (auxiliary task) and a natural language specification of the target object. Our model combines indirectly specified object locations with labeled object identifications and reach-for-grasp movements (illustrated in \autoref{fig:setup}).

In addition to extending the model for performing a three-dimensional visuomotor task and developing a novel data set, our main contribution is a detailed analysis of the model: We evaluate for systematic biases in detecting objects based on the distribution of different color and shape combinations, as well as the spatial distribution of objects in the dataset. We compare the achieved detection accuracy for shape color combination and at different positions. Furthermore, we provide a better insight into the effect of training with auxiliary tasks in an integrated visuomotor model by training different combinations of tasks simultaneously. We can show that the primary visuomotor task can be successfully trained simultaneously with one of the two auxiliary tasks. Code and datasets are  publicly available\footnote{\url{http://knowledge-technology.info/research/software\#multimodal_multitask_grasp}}.

\section{RELATED AND PREVIOUS WORK}
In \autoref{sota:objectDetect}, we provide an overview of the Retinanet model, its Resnet backbone, and the feature pyramid network for object classification and localization, which we extended to accommodate our visuomotor task. In \autoref{sota:grasping}, we present relevant related neural approaches for visuomotor learning and in \autoref{sota:multitask}, the related work on multi-task learning and training with auxiliary tasks is reported. In \autoref{sota:previous}, we present details on relevant previous work.

\subsection{Object Detection with Neural Vision Networks}
\label{sota:objectDetect}
Object detection is the combined problem of classifying and localizing an object in an image. For classification, a Convolutional Neural Network (CNN) is an established solution. Such models have a conic structure of interleaved convolutional and pooling layers. As deeper layers represent higher-level features at a lower resolution, CNNs perform well at determining \emph{what} is shown in an image but not precisely \emph{where} it appears.
In contrast, models for object localization rely on accurate spatial information to detect the position of an object. Two-stage models utilize two independent neural networks. The first stage proposes a set of regions in the input image that are likely to contain objects, whereas the second stage classifies those candidate regions~\cite{girshick2014rich}. The underlying assumption is that objects that completely fill the proposed region will be classified with high confidence.

Single-stage models were introduced to reduce the number of necessary computations. 
In single-stage models, both the proposal generation and classification share the same convolutional features extracted by a common subnetwork often referred to as a backbone~\cite{ren2015faster}. 
Single-stage approaches such as YOLO~\cite{redmon2016you} require less processing time by classifying objects over a regular sampling of possible locations. Until the release of Retinanet~\cite{lin2017focal}, single-stage models were outperformed by two-stage models. 
The Retinanet has been shown to outperform other single-stage~\cite{redmon2016you, liu2016ssd, fu2017dssd} and multiple stage detectors~\cite{girshick2014rich, ren2015faster}, for instance, on the challenging COCO~\cite{lin2014microsoft} dataset. This success is partially attributed to the \emph{focal loss}, employed for coping with the class imbalance in the training data, which features an overwhelming number of easy-to-classify negative samples. These \emph{easy negatives} show only small parts of an object or the background. While a simple factor can give more weight to positive samples, the focal loss also de-emphasizes the loss of well-classified samples while enhancing the loss of misclassified samples. The focal loss ensures that the gradient for updating the network is dominated by those samples that are still difficult to classify and not those that the network can already classify well, which would lead to a stagnation of the learning process. The focal loss is defined as:
\begin{equation}
    \textrm{\textit{FL}}_{class.}(p_t) = -\alpha(1 - p_t)^\gamma log(p_t)
\end{equation}
where $p_t \in [0,1]$ is the model's probability estimate for a given class, $\alpha \in [0,1]$ is a balancing factor for weighing the influence of a class, and $\gamma \geqslant 0$ defined as the tunable focusing parameter which exponentially down-weighs easy examples.
The Retinanet model consists of a Feature Pyramid Network (FPN) and a deep residual network called a Resnet, forming its backbone. 
The residual network is a convolutional network with skip connections; these are connections that propagate the error across layers, thus assisting in overcoming the vanishing gradient problem, particularly for deep models. 
The Resnet is connected at multiple stages to an FPN. An FPN can be seen as a complementary top-down path to a CNN, utilizing lateral connections to build a high-level semantic feature map at different scales. 
The FPN overcomes the CNN's limitation of having layers with either high-level features or high spatial resolution. 
Regression and classification subnetworks extend from the FPN. The detection output (classification and localization) is extracted from the FPN heads.

\subsection{Neural Approaches for Visually-guided Grasping}
\label{sota:grasping}
Conventional approaches for visuomotor applications like vision-based grasping are usually modular. Each module is independent and expert-designed, e.g.,~\cite{leitner2016modular}. Developmental robotics takes a different approach, relying on emerging abilities and learning of complex skills through interaction with the environment, in contrast to hand-crafted or hard-coded systems~\cite{cangelosi2015developmental}. In developmental robotics, computational models and learning setups are still designed by experts. However, these models learn and adapt over time.
The adaptation is inspired by findings from biological systems. Humans are born with an existing brain structure; however, interaction with the environment is necessary to develop complex visuomotor and cognitive abilities. In state-of-the-art approaches, this learning can be realized with artificial neural networks. Approaches that rely on deep reinforcement learning~\cite{lillicrap2015continuous} require many samples; the resulting training times are often not suitable for physical robot platforms. Therefore, approaches to increase the sample efficiency of (continuous) deep reinforcement learning have focused on sample selection strategies~\cite{schaul2015prioritized} or biologically inspired methods for accelerating the learning process, like curiosity-driven exploration~\cite{hafez2017curiosity, MikhailCuriosityDriven} as well as curriculum and incremental learning~\cite{Eppe2019_CGM, NolfiIncementalLearning}. A faster approach is to transform the reinforcement learning task into a supervised learning problem by generating fully annotated training samples of visuomotor actions~\cite{levine2016end}. Kerzel and Wermter~\cite{kerzel2017neural} let a robot place an object at random positions on a table, reversing the target grasping problem to an object placement task instead. Compared to reinforcement learning approaches, the robot's physical interaction with the environment is reduced but still required. Learning with auxiliary tasks could be another method to enhance the sample efficiency for visuomotor learning approaches.

\subsection{Multi-task Learning and Training with Auxiliary Tasks}
\label{sota:multitask}
Expanding the range of tasks learned by a single model, often called ``learning with auxiliary tasks" or ``multi-task learning", has shown to improve a model’s performance in the primary task and the auxiliary tasks~\cite{caruana1997multitask} alike. Auxiliary tasks can be crucial to the final objective~\cite{gornitz2011hierarchical} or integrated purely as a form of regularization~\cite{sanabria2018hierarchical}.

\begin{figure*}[ht]
\centering
  \includegraphics[width=1.0\linewidth]{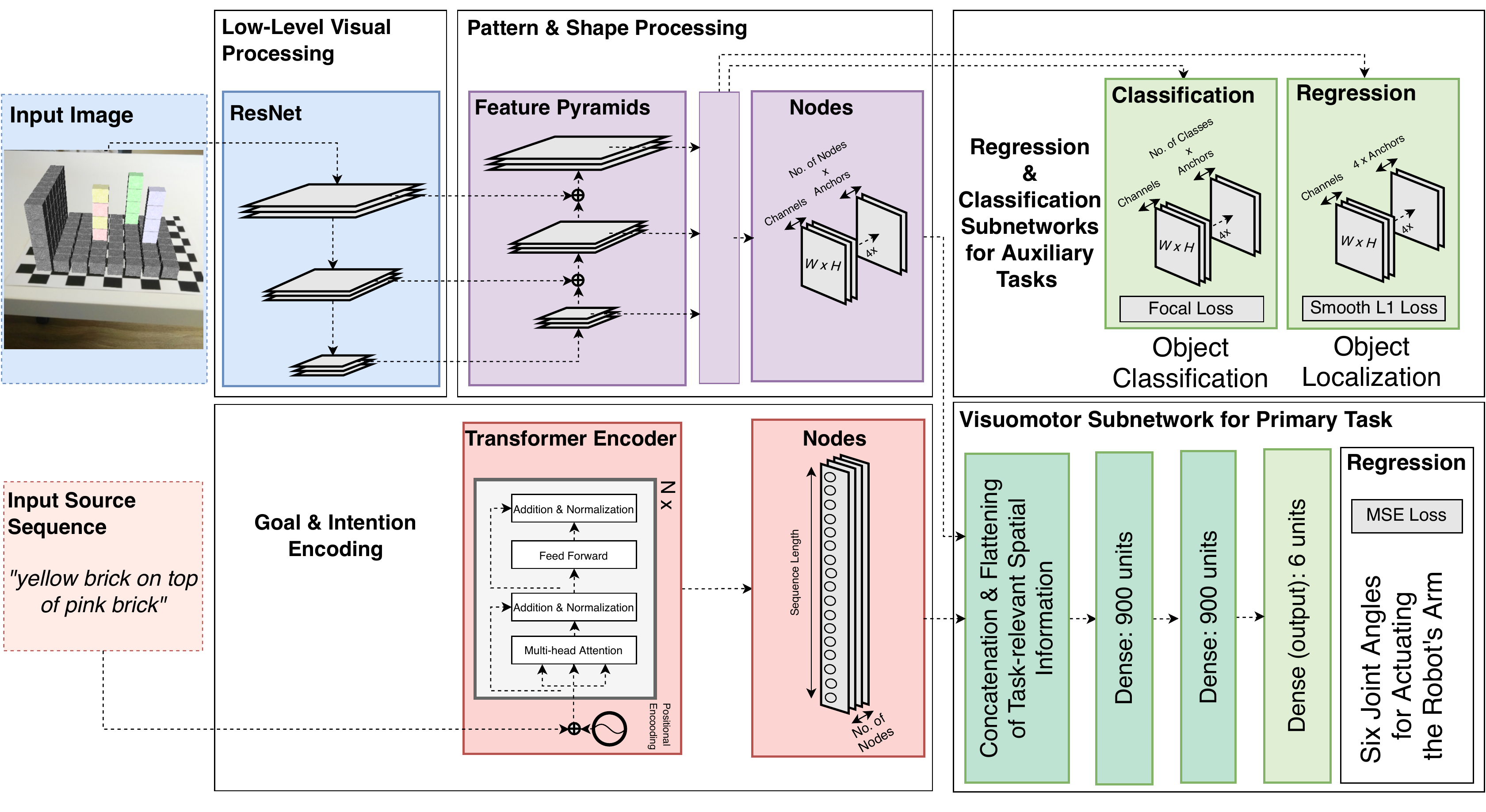}
  \caption{Extended visuomotor model based on~\cite{self}. A shared visual processing is used for the primary visuomotor and two auxiliary vision tasks. A Transformer Encoder encodes goal information for usage in the primary task.}
  \label{fig:fullmodel}\vspace{\squeezefigs}
\end{figure*}

Multi-task learning approaches are often realized with multi-output neural models. A single task can potentially overfit to specific patterns; however, introducing more tasks can lead to an averaging of the noise across them, making the model more robust to irregularities. In multi-task neural models, the early-layer parameters can be shared between tasks. Each task also has a number of separate layers (unshared subnetwork) with dedicated parameters that are not shared. The output of the neural network is represented by a combination of all learned tasks. These tasks can have different objectives with various loss functions. The overall loss is a weighted summation of all the losses across the tasks. The overall (combined) loss is the total of the weighted losses, yet the error propagates from each output layer independently. The error does not leak from one unshared subnetwork to another, backpropagating through each unshared subnetwork depending on its associated loss and any preceding (shared) layers.

A non-trivial problem arising from the different losses introduced to a single neural network is known as destructive interference~\cite{zhao2018modulation}. Destructive interference refers to the effect of tasks driving the gradients in opposing directions during backpropagation, as the weights are being adjusted for the layers shared across all tasks. Zhao et al.~\cite{zhao2018modulation} proposed a modulation module which applies task-specific masks to layers within the network. The masks reduce the gradient angles across tasks internally since they are integrated as learnable parameters within the layer, mitigating destructive interference. However, the best approaches in multi-task learning are an open research problem and are yet to be established. Although many approaches have been proposed to overcome common issues in multi-task learning, such approaches remain task-specific. 
However, tuning the tasks and integrating multiple models can, in some cases, improve a neural network’s performance~\cite{kaiser2017one}.

From a biological perspective, humans learn to perform complex tasks by integrating knowledge acquired through performing simpler tasks. This is similar to learning with auxiliary tasks: multiple simple auxiliary tasks are learned, which, in turn, help the learner achieve the primary objective. It is likely that the development of visual object representation and visuomotor abilities are linked~\cite{HKSW18}. 

\subsection{Previous Work on Shared Visuomotor Model}
\label{sota:previous}
As we present a follow-up study on our previous work addressing a neurocognitive-inspired visuomotor model for a robotic reach-for-grasp \cite{self}, we will summarize the experimental setup and results to explain and motivate our follow-up study. The underlying neural model is based on Retinanet \cite{lin2017focal} and a neural end-to-end approach for visuomotor learning by environment interaction \cite{kerzel2017neural}. The model was both trained and successfully tested on a physical robot, revealing an object-specific bias that we attributed to the training set. A detailed description of the original dataset, the non-extended shared visuomotor model, and the optimization of the relevant hyperparameters for training can be found in~\cite{self}.

\subsubsection{NICO Robot, Dataset and Physical Experimental Setup}
The neurocognitive visuomotor model is embodied in the 
developmental robotic platform NICO, the Neuro Inspired COmpanion~\cite{Kerzel2017NICO} (see~\autoref{fig:setup})\footnote{ Further information and videos: \url{http://nico.knowledge-technology.info}}.
NICO's anthropomorphic design is based on the body proportions of a  child at four to five years of age. With a height of about one meter, it can interact with domestic environments while having a non-threatening appearance and maintaining a light weight. NICO's arms have six Degrees of Freedom (DoF) and resemble the human anatomy: three DoFs in the shoulder, one DoF in the elbow, and two DoFs rotate and flex the hand. NICO is equipped with two three-fingered Seed Robotics SR-DH4D\footnote{ \url{http://www.seedrobotics.com}} hands that are tendon-operated and capable of gripping small objects. Two cameras are fitted in NICO's head.


In the physical experimental setup, NICO sits at a table and looks down at a set of reachable objects, as shown in \autoref{fig:setup}. Kerzel et al.~\cite{self} created the dataset in two steps: \textbf{1)} Using a self-learning paradigm~\cite{kerzel2017neural, Eppe2017} to create a dataset for single object grasping; \textbf{2)} Augmenting the images with distractor objects via image manipulation. In total, 232 training samples, equally distributed over five different training objects, were created. Each sample consisted of an image from the robot's perspective showing the target object and a number of distractor objects, the joint configuration to reach for the target object, and semantic information describing the target object.

\subsubsection{Object Picking with and without Distractor Objects}
The model was initially trained to reach for a target object in a scene without distractor objects or semantic information.
All experiments were repeated ten times with randomly initialized weights. The MSE is computed over the validation set (10\% of the dataset); it averaged to 0.002. Adding semantic information about the object to be grasped did not cause significant changes. The results are interpreted as evidence that the end-to-end learning of the grasping task is possible with the model.

Next, each image was augmented with one to three distractor objects. An n-hot encoding of object properties (type, color, shape) was used to select the target object. 
Results showed that the MSE averaged over ten training trials on the validation set (10\% of the dataset) increases with the number of distractor objects in the scene. The MSE increases from 0.0006 (one distractor object) to 0.0021 (two distractor objects) and 0.0051 (three distractor objects). To analyze the exact nature of this performance decrease, Kerzel et al. embodied the model into the physical robot for grasp experiments.

\subsubsection{Robotic Experiments}
The embodiment of the visuomotor model into a physical robot allowed us to evaluate it under realistic conditions and to directly observe the robot's visuomotor behavior for a more detailed analysis, i.e., if the increase in the MSE caused by additional distractor objects reflects an overall decrease in grasp accuracy or grasping non-intended objects. A grasp is counted as successful if the targeted object is reached (touched) by the robot's hand and could be lifted completely after enclosing it with the fingers.

To establish a baseline, the robot grasps a single object in a scene without distractor objects. The object is placed into each position of a 3~×~6 grid in the 30~×~60 cm workspace. With six different objects, an average grasp accuracy of about 96\% was achieved, which exceeded the accuracy of earlier models~\cite{kerzel2017neural}. This performance increase was attributed to the pyramidal vision model's ability to preserve spatial information.

Next, the model's ability to pick an object in the presence of one to two distractor objects was evaluated. The target object was placed in the above-described workspace and grid, while the distractor object(s) shifted their position accordingly to be non-connected and non-overlapping with the target object. Overall, 36 grasp attempts with different objects were analyzed. In line with the observation on the MSE from the previous experiment, distractor objects decrease the grasp accuracy to about 67\% and 50\% for two and three objects on average. A frequent reason for failed grasps were small deviations from an optimal grasp configuration. However, a tendency to grasp a distractor object could also be observed, accounting for $\sim$20\% of all failed grasps. Therefore, a significant part of the error increase in the presence of distractor objects can be attributed to the model's difficulty in correctly identifying or localizing the target object. This hypothesis is further supported by the observation that grasping the wrong object occurred more frequently when objects were visually similar. The model's issues with correct object location could stem from biases introduced by the relatively small training set or the pre-trained vision models. To address this issue and allow an automated and more thorough analysis, we designed the presented follow-up study.

\section{NEURAL MODEL AND METHODOLOGY}

We first present the extended visuomotor model based on~\cite{self}. We describe the shared and independent subnetworks of the model for realizing the primary task of grasping a targeted object, as well as the two auxiliary tasks of localizing and identifying objects. 
We provide a detailed description of the realization of the experimental setup and the new dataset generated with the help of augmented reality. The dataset annotations are acquired from the Extended Train Robot (ETR) dataset by Alomari and Dukes~\cite{ETR}.

\subsection{Neural Shared Visuomotor Model}

The shared visuomotor model enables the simultaneous learning of object classification and localization, as well as motor control. The network is designed to receive sensory input in the form of an image displaying a table with one or more objects, viewed from the robot's perspective, as well as a neurally encoded description of the grasping target. Our model has three output layers:
\textbf{1)} It generates robotic arm joint angles for performing a reach-for-grasp action as a primary task; \textbf{2)} It generates bounding boxes for the auxiliary task of localizing objects in the image; \textbf{3)} As a second auxiliary task, it classifies the localized objects. The two auxiliary tasks of object localization and classification can be jointly evaluated as an object detection task to analyze the model's ability for visual (but not motor) processing. The network can be trained on the primary visuomotor task only or in combination with any number of auxiliary tasks.

We extend the model developed by Kerzel et al.~\cite{self} by introducing two auxiliary tasks and redesigning the object encoding mechanism, as shown in \autoref{fig:fullmodel}. The model is based on the Retinanet~\cite{lin2017focal}, built upon a Keras-based\footnote{\url{https://github.com/fizyr/keras-retinanet}} implementation of the Retinanet using a Resnet~\cite{he2016deep} with 50 layers as a backbone. The Retinanet extends the backbone with lateral connections to the FPN. The FPN infiltrates the backbone at three stages, with two additional convolutional layers as described in~\cite{lin2017focal}, resulting in a total of five pyramidal layers. Each pyramidal layer has 256 output channels. The Resnet mimics the early hierarchical processing of visual input in the human brain~\cite{kruger2013deep} while the feature pyramids contain pattern and shape information with regard to regions in the input image. This shared visual processing is then used for the primary visuomotor and the auxiliary object classification and localization tasks by different subnetworks.

The input comprises a 990~×~540 pixel RGB image and a neural encoding, describing the target object in natural language sentences, e.g., \emph{``the purple pyramid on top of the blue and green stack of blocks"}. The neural encoding for a given sentence is generated by a Transformer Encoder~\cite{vaswani2017attention}. 
The transition to a three-dimensional world necessitates the use of more complex object descriptions that contain not only visual features but also descriptions of object positions since blocks of the same shape and color can appear in multiple locations. For the remainder of the text, the Transformer Encoder can be assumed to create a unique neural representation of the target object, analogous to the one-hot encoding used by Kerzel et al.~\cite{self}. Due to the increased complexity of object description in terms of location and shape in a three-dimensional world (e.g., \textit{``yellow brick on top of pink brick"} as shown in~\autoref{fig:fullmodel}), a simple one-hot encoding would not suffice; hence, the employment of the Transformer Encoder. The Transformer utilizes dot-product attention to prioritize words that contribute most to the end-goal. More specifically, the Transformer Encoder learns a language model for encoding natural language sentences, which makes it suitable for our task: describing the object to be grasped and allowing the neural model to attend to relevant words that fulfill our desired objective. The output encoding is presented in the form of a dense representation.

Three parallel subnetworks branch from the Feature Pyramids, two of which are part of the original Retinanet setup that will be used for training the auxiliary tasks, and one for enabling the primary visuomotor task. The classification subnetwork which predicts the probabilities of objects existing at each position and the regression subnetwork for estimating the limits of the bounding boxes. Following the Retinanet implementation \cite{lin2017focal}, actually, multiple classification and regression subnetworks branch off different layers of the FPN; this is simplified to enhance clarity. The visuomotor subnetwork, in which the information from the pyramidal network is combined with the encoded object description to produce joint angles.
The model's model is shown in \autoref{fig:fullmodel}. Parameters are not shared across the subnetworks, which are designed as follows: 

\emph{The classification subnetwork} is constructed using four 3~×~3 convolutional layers with 256 channels for each FPN layer, and parameters shared across all the FPN layers. Each convolutional layer uses a ReLU activation, with a sigmoid activation for the final layer. The outputs are of size \textit{K~×~A}, where \textit{K} represents the number of object classes to be detected, and \textit{A} represents the number of anchors. Anchors are windows of predefined shape and size that slide over the input image according to a stride parameter. As the predetermined size of the anchor can be an issue in detecting arbitrary objects, both the number and shape of the anchors were empirically optimized for our approach. We use nine anchors for all experiments. The final layer results in an output indicating whether a class belongs to an anchor or not, along with the class index. Focal loss is the cost function used for this subnetwork.

\emph{The regression subnetwork} is constructed using four 3~×~3 convolutional layers with 256 channels for each FPN layer, and parameters shared across all the FPN layers. Each convolutional layer is followed by a ReLU activation, with a sigmoid activation for the final layer. The network outputs the location (x, y) and size (height, width) of each anchor. The four linear outputs per anchor are regressed to the nearest ground-truth box. The smooth L1 loss is the cost function used for this subnetwork.

\emph{The visuomotor subnetwork} combines the semantic encoding of the targeted object with the visual features from the Feature Pyramids that represent the scene. Extending our previous work, we introduce an intermediate, multi-dimensional structure between the visuomotor subnetwork and the low-level visual processing module. We refer to this structure as a node, with multiple nodes indicating the dimensionality. The intermediate structure connecting the visuomotor subnetwork with the Retinanet is similar to the regression subnetwork. We set the number of nodes to four, resembling the output of the Retinanet's bounding box regression subnetwork. We introduce the nodes to reshape the output branching from each module and integrate seamlessly with their pre-existing design. Similarly, we introduce semantic encoding nodes feeding from the goal and intention encoding module. These nodes are composed of a dense layer with a length of 200 units. We set the number of nodes to four for maintaining consistency with the visual processing nodes. The visuomotor subnetwork flattens and concatenates the visual processing and semantic encoding nodes. The concatenated layer is followed by two dense layers with 900 units each. The two dense layers use a sigmoid activation function, followed by the final output layer, with six units representing the six-degrees-of-freedom of the robotic arm. The six units produce the joint configuration used for actuating the arm and grasping the target object. The layers following the concatenation learn the shared features between the two modules and are based on the post-convolutional layer structure described in~\cite{kerzel2017neural}. The Mean Squared Error (MSE) loss is used as this subnetwork's cost function.

\subsection{Novel Augmented Reality Dataset}
We create a novel augmented reality dataset for a three-dimensional reach-for-grasp task. By using augmented reality and a simulation environment, the dataset is created with all annotations for training and evaluating our model on the primary as well as the auxiliary tasks. Each sample consists of an image of an arrangement of geometric objects, object classes in terms of shape and color, object locations in terms of bounding boxes, and, most importantly, joint configurations to grasp these objects. The reason for creating a novel dataset instead of extending the dataset from our previous work is the change of the task from reaching for an object on a planar surface to objects in a three-dimensional space. Recording such a dataset is time-consuming, which limits the number of generated samples.
As a controllable environment for the dataset, we chose a three-dimensional block world scenario. This extends the existing dataset with a more challenging task: the environment is no longer limited to salient and sparsely spread objects but contains smaller and potentially occluded targets, which are stacked on top of one another. In this three-dimensional block world, blocks with basic shapes can be picked from different locations. Our dataset is based on the \emph{Extended Train Robots} (ETR) dataset~\cite{ETR}, which offers an intuitive setup of stacked objects. Additionally, the target object is not just described by shape and color but by their spatial position in the image. This is vital for scenes with multiple identical objects. The descriptions are variable in size depending on the complexity of the scene and can be neurally encoded via the Transformer Encoder.
While the ETR provides the spatial layouts of the blocks and linguistic instructions, it lacks the necessary annotations for our primary and auxiliary tasks. We extend the dataset in three steps: \textbf{1)} Augmented reality is utilized to construct visualizations of the block world layouts, that are well suited as input to the shared visuomotor network; \textbf{2)} Information about the block layout and the augmented reality creation process is used to construct annotations for training the two auxiliary tasks of object localization and classification; \textbf{3)} A simulation environment and an inverse kinematics solver are used to estimate the joint configurations for grasping different blocks to train the model on the primary task.

\begin{figure*}[t]
  \centering
  \includegraphics[width=1.0\linewidth]{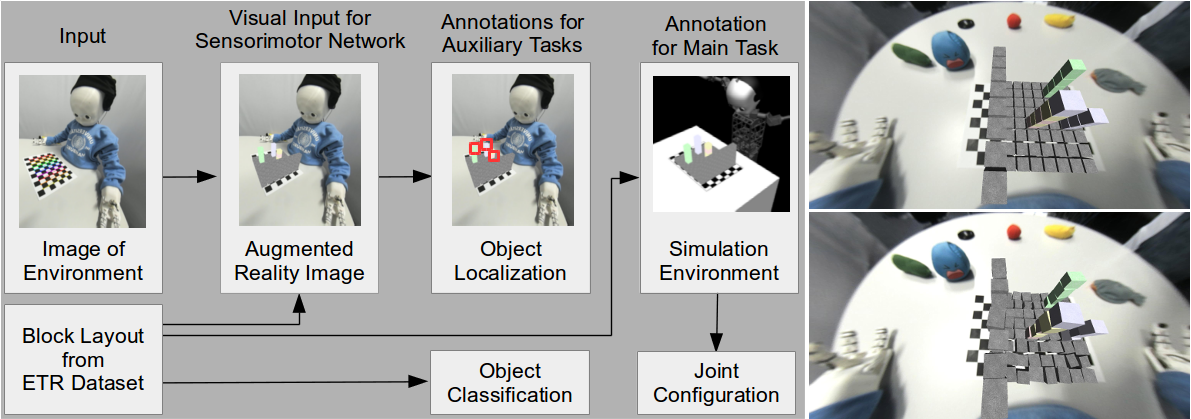}
  \caption{Left: Generation of the novel dataset. Based on a background image and a block world layout from the ETR dataset, an augmented reality image of a layout is created as input for the shared visuomotor network. Information from the layout and the augmented reality creation process is used to extract annotations about object classification and localization for the training of the auxiliary tasks. Finally, the scene is recreated in a simulation environment where an inverse kinematics solver is used to generate joint values for grasping the object to train the primary task. Right: Examples of generated augmented reality images from the robot's perspective for training without and with noise applied to object positions. The environment shows a desk with distractor objects.}
  \label{fig:datasetGen}\vspace{\squeezefigs}
\end{figure*}

\subsubsection{The Extended Train Robots Dataset}
The ETR is a synthetic dataset published by the University of Leeds~\cite{ETR}. The dataset is a collection of 625 layouts in an 8~×~8~×~8 block world, with natural linguistic instructions describing objects and spatial interactions~\cite{Dukes2013}. Non-expert users, employed through the Amazon Mechanical Turk platform, were asked to annotate commands for 1000 scene pairs. Each scene pair contains the initial and final configuration for achieving the requested goal. These configurations include the block layouts and positions of the end-effector. The ETR layouts describe the blocks and their locations for a given scene. The blocks come in two different shapes, cubes and pyramids, with the following colors: yellow, white, gray, magenta, blue, cyan, red, and green.
In our shared visuomotor model, we tokenize the words derived from the natural linguistic instructions, splitting them by the spaces and feeding them into a language encoding Transformer Encoder. Example descriptions of target objects in the ETR dataset include: \emph{the purple pyramid on top of the blue and green stack of blocks} or \emph{the turquoise pyramid sitting on top of the green and blue slab}. These descriptions are combined with action descriptions like \emph{remove} or \emph{pick}.
Although the ETR's linguistic data provides information about the block relocation, we are only interested in simply picking the specified objects and not releasing them elsewhere. The linguistic data (command sentences) is yet essential, since it offers the necessary unique description of a targeted block, given that multiple blocks with the same color and shape might appear in a single layout. These sentences also provide knowledge about the approximate position of the blocks. This helps to disambiguate the targeted object. During preprocessing, we discard all language descriptions and layouts, which are marked as faulty by the ETR dataset authors.

\subsubsection{Synthetic Visual Dataset Augmentation}
To generate the visual input data for the visuomotor network, we create synthetic images by rendering the ETR block world layout on top of real images, captured in the robot's laboratory environment; the process is shown in ~\autoref{fig:datasetGen} (left). To ensure correct alignment, a checkerboard is used for calibration; any distortions or fisheye effects in the real images caused by the camera lens are removed using OpenCV~\cite{opencv_library}.

We capture multiple images showing the checkerboard sheet lying on a table with different laboratory backgrounds from the robot's perspective. We set the aspect ratio to 16:9, resulting in images with a width and height of 1920~×~1080 pixels. Fifteen different backgrounds were chosen for the dataset. The backgrounds show the robot's laboratory environment from different positions. The captured images vary in background, lighting condition, table surface, and presence of distractor objects. By changing the background using different capturing sites, we minimize the influence of the background. Different light settings in the laboratory were used to vary the illumination. Finally, various distractor objects in the form of toys were randomly placed within the camera’s field of view. The distractor objects are different from computer-generated cubes and pyramids; they provide negative examples and assist in regularizing the dataset.

We selected 12 background images (3 with distractor objects and 9 without) for training, with the remaining three images for validation. The three validation backgrounds were the closest to the robot's perspective, ensuring the robot's static pose relative to the grasping region (defined by the checkerboard's position). One of the validation backgrounds contained distractor objects. We did not apply cross-validation, since the physical robot would only capture images from its own perspective during the inference phase, meaning that the 12 varying background perspectives are only introduced for augmenting (as a form of regularizing) the training dataset. Although we do not deploy the model to a physical robot, we validate our results under the assumption that such an approach could potentially be applied to a real-world scenario.

We superimpose three-dimensional computer-generated objects onto the background images using the estimated pose of the checkerboard in each image. Pose estimation is performed with a random sample consensus~\cite{fischler1981random} to solve the perspective-n-point problem. The object positions and descriptions are extracted from the ETR dataset. We then utilize the OpenGL~\cite{shreiner1999opengl} library for creating three-dimensional blocks with their respective colors. We map the discrete block positions to the checkerboard pattern lying atop the table. Using the extrinsic camera parameters, the blocks are re-scaled and rotated to match their estimated pose in the camera view. During this process, we add a small amount of noise to the object position and rotation for generating realistic and feature-rich images and facilitating robust learning. \autoref{fig:datasetGen} (right) shows examples of a generated augmented reality image from the robot's perspective with \autoref{fig:datasetGen} (bottom) and without \autoref{fig:datasetGen} (top) noise applied to object positions and rotations. During the augmentation process, we acquire the bounding boxes surrounding each block in view. The bounding box coordinates are used as ground-truth data for training the regression (localization) subnetwork.

To ensure robust learning, we apply noise during the three-dimensional augmentation process. The blocks' scales are randomly varied in all dimensions within $10\%$ of their original size. They are also randomly rotated within $6^{\circ}$ on the x and y axes, and $10^{\circ}$ on the z-axis, relative to the block’s centroid. Finally, the blocks are displaced by up to 1 cm in all directions.  \ref{fig:datasetGen} (right) shows an example of a block layout with noise. The contrast and lightness of the background, as well as the virtual OpenGL light source, are varied.

We examine the block location distribution by counting the blocks appearing in each discrete position. Since blocks must be placed on top of each other, fewer blocks will be observed higher up the stack. We, therefore, ignore the z-axis. As seen in \autoref{fig:posBias}, the density of the block distribution tends toward the center and the corners of the grid. Although the region with the highest occurrence of blocks has about five times more objects than the region with the lowest density, we observe that the blocks are distributed and not clustered in a few regions.

\subsubsection{Joint Coordinate Data from Simulation Environment}
The ETR dataset provides us with the discrete Cartesian coordinates for all blocks in a three-dimensional block world space. For training the shared visuomotor network, we require the configuration of the robot's arm for reaching these positions. As the process of acquiring the arm positions in the real world would be prohibitively time-consuming, we recreate the scene in a simulated environment instead. The simulated robot model is described by the Unified Robot Description Format (URDF), containing the kinematic and visual specifications of the robot. We convert an already existing URDF model of NICO to a format compatible with the Multi-Joint dynamics with Contact (MuJoCo)~\cite{todorov2012mujoco} simulator. The MuJoCo simulator is a model-based physics engine, which specializes in simulating robots designed for industrial and research purposes.

\begin{figure}[t]
  \centering
  \includegraphics[width=1.\linewidth]{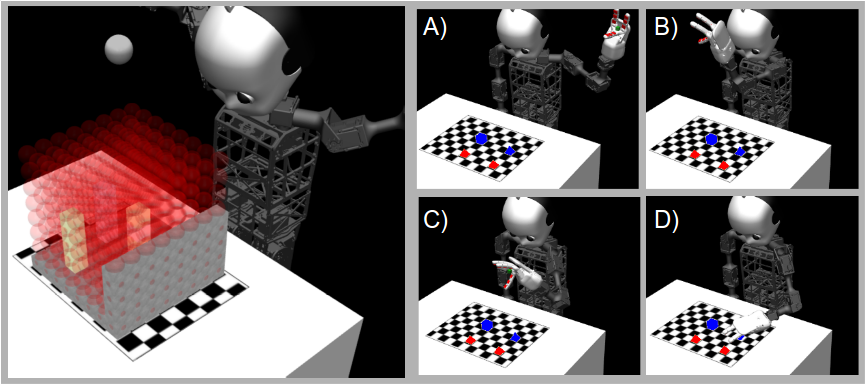}
  \caption{Realization of the grasping scenario with NICO in the MuJoCo simulation environment. Left: Translucent spheres mark the possible positions for blocks. Right: The four steps for grasping a block. NICO prepares for grasping by lifting its hand (A), moves the hand above the table (B), lowers it towards the table surface (C) before grasping the block (D).}
  \label{fig:datasetSim}\vspace{\squeezefigs}
\end{figure}

\autoref{fig:datasetSim} shows the simulation environment, with NICO seated at a table onto which block layouts from the ETR dataset are placed. The visual properties of the simulation are not relevant here since the more realistic augmented reality images are used as input to the shared visuomotor network. However, the positioning of the robot in relation to the blocks needs to be precise to generate accurate training data.
Similar to the augmented reality setup, the blocks from the ETR dataset layouts are placed above the square spaces of the checkerboard pattern. To reach the blocks using the robot's arm, we need to acquire joint angles that facilitate the movement of the gripper. The joints of interest are those which describe the robot arm's pose. We use NICO's left arm for grasping objects and are interested in acquiring its six joint angles. We create an 8~×~8~×~8 grid for positioning the blocks in predetermined locations, as shown in \autoref{fig:datasetSim} (left). We use MuJoCo's Newton solver to acquire NICO's arm configuration as we direct the end-effector of its gripper towards the targeted object's position in space. 

To ensure regularity and physical plausibility in NICO's motion as it reaches for the blocks, we enforce a sequence of actions, as shown in \autoref{fig:datasetSim} (right). The robot initially prepares for grasping by lifting its arm. It then positions the hand above the table and lowers it towards the surface, before grasping the block. Once the robot's hand reaches the block, the actuator angles are recorded. The angles are stored with their corresponding block and layout identifiers.

\subsubsection{Summary of Augmented Reality Dataset}
The novel dataset includes a total of 625 block layouts with 15 different backgrounds. For each combination of layout and backgrounds, the synthetic images, along with three noisy versions (noise applied to blocks), are created, resulting in a total of 37,500 fully annotated training samples. Images are resized to 990~×~540 pixels. For each image, one block is selected for grasping. The block position in terms of a bounding box, its class in terms of shape and color, as well as the joint configuration for reaching it are all available as part of the generated dataset. In comparison to the real-world dataset used by Kerzel et al.~\cite{self}, the current dataset is larger by a factor of $\sim$100. The increase in size compensates for the complexity introduced by having to grasp objects in a three-dimensional layout, compared to objects lying flat on a table. 

\section{EXPERIMENTS AND RESULTS}
\label{sect:experiments}
In \autoref{sect:exp_first}, we focus on the auxiliary tasks of object classification and localization combined in the task of object detection to optimize the network's vision backbone, and to evaluate if there is a systematic bias towards specific visual properties or regions in the visual processing of the model, i.e., if the visual model can more accurately detect objects of a certain shape or color or in certain regions, which could in turn influence the visuomotor task.

In \autoref{sect:ablation}, we evaluate the reach-for-grasp accuracy of the model and the effect of multi-task training on the learning of visuomotor abilities. We present a combinatorial study in which different combinations of auxiliary tasks are trained alongside the main visuomotor task.

\subsection{Selection of Network Backbone and Evaluation of Auxiliary Tasks of Object Classification and Localization}
\label{sect:exp_first}

We optimize the model by selecting the best-performing vision backbone. We then analyze if there is an inherent bias towards particular objects or locations in the visual processing of our model. We evaluate the object detection (a combination of the localization and classification auxiliary tasks) performance using the mean Average Precision (mAP). The mAP is used to analyze for a systematic bias in recognizing objects with certain visual properties or at certain positions. An object is counted as correctly detected if its bounding box has an intersection-over-union (IoU) $\ge$ 0.5 with the ground truth and is and correctly classified in terms of shape and color. We follow the approach presented in~\cite{everingham2010pascal} for calculating the mAP on evaluating the object detection performance. In our previous work \cite{self}, we hypothesized that grasp errors could mostly be attributed to issues in the visual processing stream. In contrast to our previous work, we can now directly observe whether a systematic bias exists in the visual processing stream, occurring due to the visual properties (shape or color) of objects or their locations.

The hyperparameters are set to match the original Retinanet implementation. We use 50k training iterations with a batch size of 1, a stochastic gradient descent optimizer with a learning rate of $0.01$, and a momentum of $0.9$. For the Focal loss we set $\alpha$=0.25, $\gamma$=2. All experiments were carried out on the synthetic dataset with an 80-20 split between training and validation data. Each experiment was repeated three times.

\subsubsection{Backbone Selection}
\begin{table}[htbp!]
\centering
\caption{The Retinanet mean average precision (mAP) with different backbones trained on the synthetic dataset over three trials.}
\label{tab:mapRetinanetBackbones}
\begin{tabular}{cc}
\hline
\textbf{Backbone Model} & \textbf{mAP} \\ \hline
Mobilenet128 & 0.124 \\ \hline
Mobilenet160 & 0.583 \\ \hline
Mobilenet224 & 0.668 \\ \hline
Densenet121 & 0.757 \\ \hline
Resnet50 & 0.872 \\ \hline
Resnet101 & 0.891\\ \hline
Resnet152 & 0.893 \\ \hline
\end{tabular}
\end{table}

All backbones used for the Retinanet are pre-trained on 1.2M images from the ImageNet\footnote{ \url{https://github.com/tensorflow/models/tree/master/research/slim\#pre-trained-models}} dataset. For all experiments, the backbones are initialized with the pre-trained ImageNet model weights. We conduct a pilot experiment on the Retinanet to determine the backbone with the best performance. We explore different backbone models with a varying number of parameters, as shown in \autoref{tab:mapRetinanetBackbones}. We compare the results achieved using Mobilenet~\cite{howard2017mobilenets} with 128,160, and 224 layers, a Densenet~\cite{iandola2014densenet} with 121 layers, and a Resnet with 50, 101, and 152 layers. Over three trials, the Resnet152 backbone achieved the best results with an mAP of 0.893. 

\subsubsection{Object Detection Accuracy}

\begin{figure*}[ht]
  \centering
  \begin{tabular}{cc}
  \subfloat{\includegraphics[width=0.45\linewidth, trim=150 50 150 85, clip]{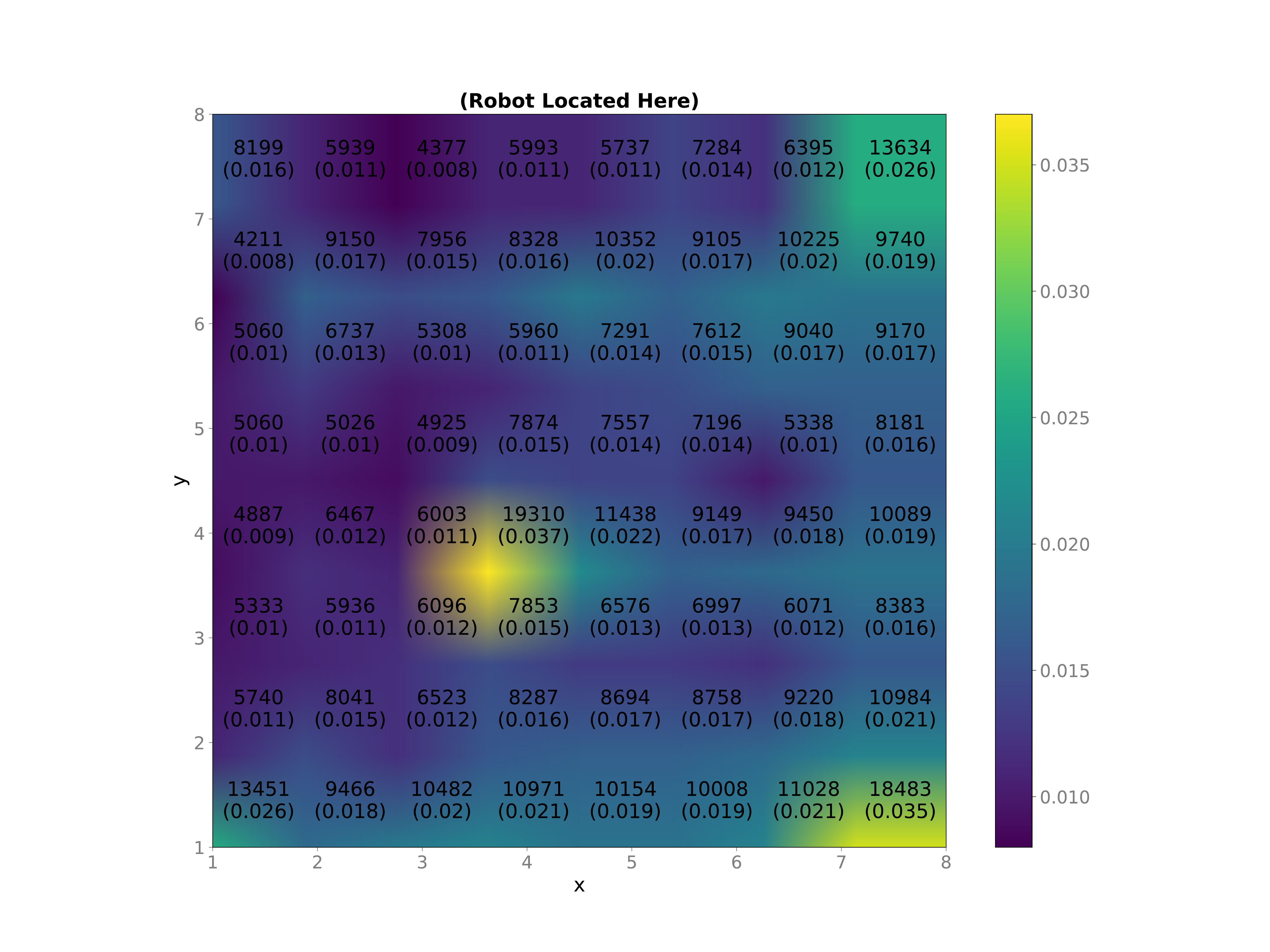}} &
    \subfloat{\includegraphics[width=0.45\linewidth, trim=150 50 150 85, clip]{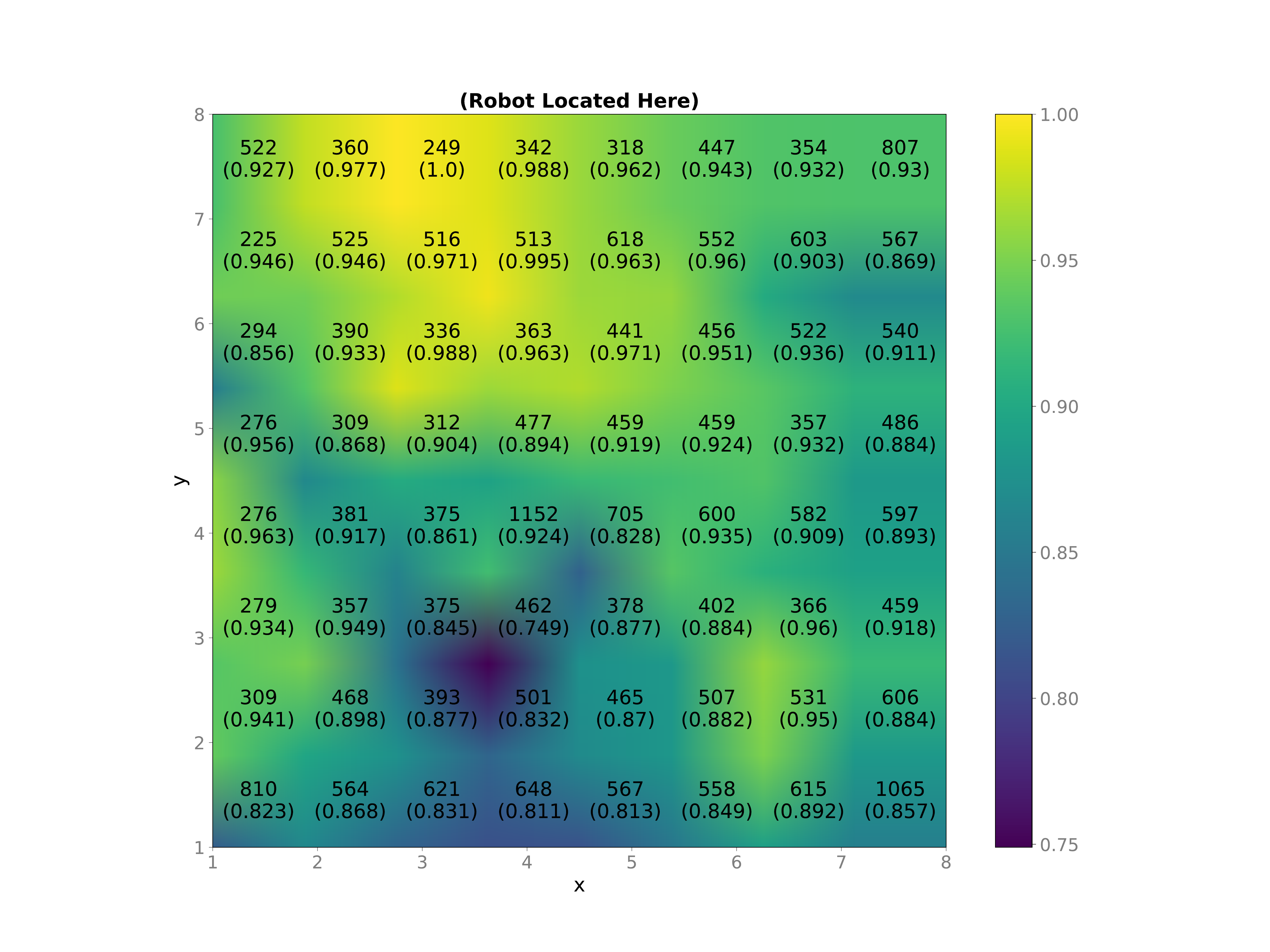}}
  \end{tabular}
  \caption{Evaluation of location bias. Position (8,8) represents the nearest leftmost block from the robot's perspective. Left: The block distribution in the combined training and validation datasets. The numbers of block instances are displayed with their normalized counts in parentheses. Right: Mean average precision for object detection of the best model trained on our dataset. The numbers of block instances are displayed with the mAP in parentheses. There is no clearly discernable mapping from high block density in the dataset (yellow areas) to areas with high accuracy on object detection (yellow areas).}
  \label{fig:posBias}\vspace{\squeezefigs}
\end{figure*} 

\begin{figure*}[ht]
  \centering
  \includegraphics[width=1.0\linewidth, trim=60 30 50 20, clip]{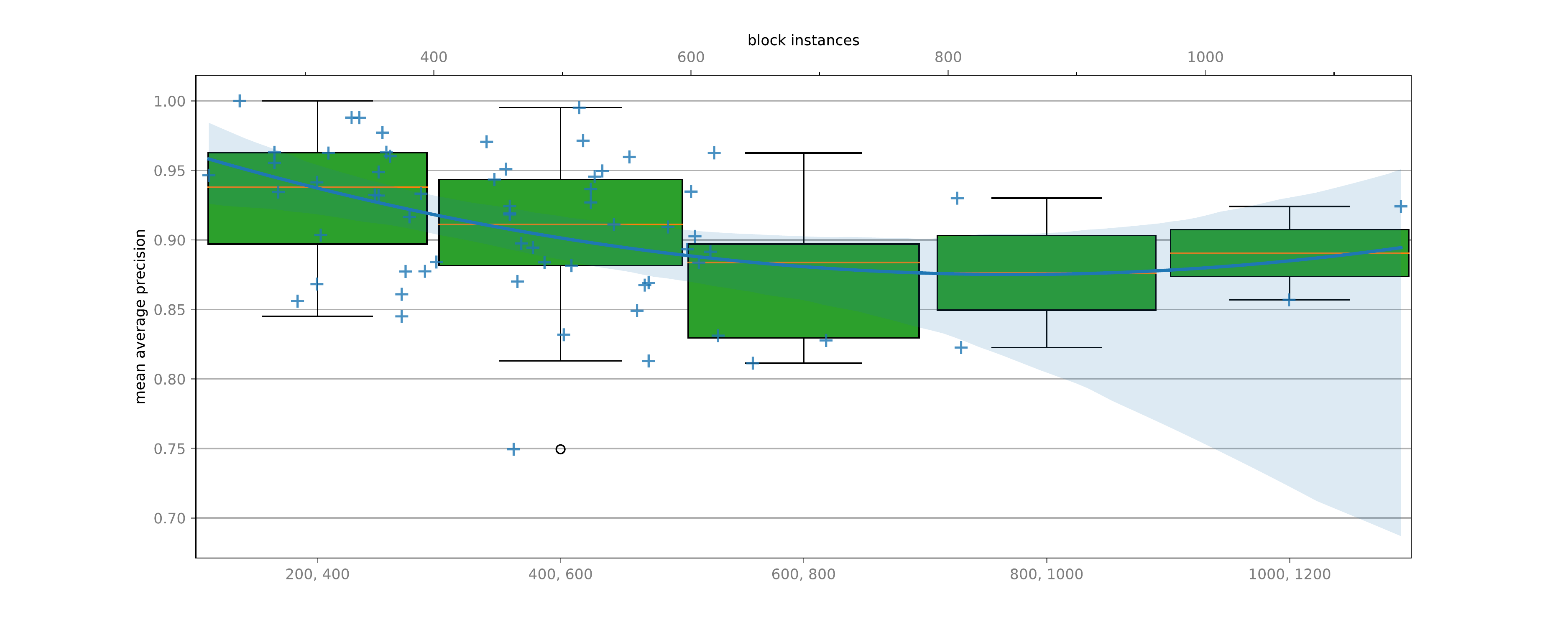}
  \caption{Evaluation of location bias in terms of mean average precision (mAP) and the number of block instances. The box plots cover the instance ranges indicated in the lower x-axis. The blue plus markers show the mAP per instance counts, whereas the blue line shows a quadratic polynomial regress and indicating a slight trend of mAP degradation with the increasing number of instances.}
  \label{fig:posBiasvsmAP}\vspace{\squeezefigs}
\end{figure*}

\begin{table}[htbp!]
\centering
\caption{Individual classes with their mean average precision (mAP) trained on the synthetic dataset.}
\label{tab:mapRetinanetResults}
\resizebox{\columnwidth}{!}{%
\begin{tabular}{c c c | c c c}
 \hline
 \textbf{Instances} & \textbf{Class} & \textbf{mAP} & \textbf{Instances} & \textbf{Class} & \textbf{mAP} \\ \hline
 297 & cube\_magenta & 0.8664 & 255 & pyramid\_magenta & 0.9545 \\ \hline
 5355 & cube\_green & 0.9143 & 342 & pyramid\_green & 0.9547 \\ \hline
 2001 & cube\_white & 0.8884 & 69 & pyramid\_white & 0.5431 \\ \hline
 6132 & cube\_red & 0.9091 & 534 & pyramid\_red & 0.9627 \\ \hline
 4128 & cube\_yellow & 0.9277 & 342 & pyramid\_yellow & 0.9611 \\ \hline
 5718 & cube\_gray & 0.8928 & 423 & pyramid\_gray & 0.8782 \\ \hline
 1125 & cube\_cyan & 0.9514 & 381 & pyramid\_cyan & 0.9585 \\ \hline
 3783 & cube\_blue & 0.9367 & 276 & pyramid\_blue & 0.7556 \\ \hline \hline 
 \multicolumn{5}{|l|}{\textbf{mean average precision}} & \multicolumn{1}{l|}{0.9130} \\ \hline
 \end{tabular}%
 }
 \end{table}

We investigate the model's capability to classify and locate objects based on their color and shape. In~\autoref{tab:mapRetinanetResults}, we report the mean average precision (mAP) results on the validation dataset for all the classes individually, based on the best Retinanet trial with a Resnet152 backbone. The Retinanet successfully classifies all objects with minor exceptions. 

Although cubes and pyramids are rarely confused, certain colors are more likely to be misclassified. We observe the worst results when it comes to white pyramids. This can be due to the permanent visibility of the checkerboard. Pyramids cover a smaller region than cubes, falling entirely within the white checkerboard boxes. This leads to empty checkerboard boxes being identified as what appears to be white pyramids. Another issue relates to the slant of the object surfaces. The pyramid surfaces facing the camera are almost perpendicular to its axis, leading to a higher reflectance than cubes. Higher reflectance results in a lighter color, making the pyramids identical to the plain white checkboard boxes. It is worth noting that all object surfaces had similar grainy textures, avoiding any latent bias that might be caused by color and texture association. The high class imbalance in the ETR dataset appears to have a negative influence on the outcomes as well. However, it is not consistently detrimental due to the focal loss.

Another factor potentially influencing our results is the location bias inherent in our dataset. We calculate the mAP across the discrete block locations instead of the classes, as shown in \autoref{fig:posBias} (right) and present the block location distribution in \autoref{fig:posBias} (left). Though a location bias in the mAP can be seen, we detect no clear pattern with regard to the distribution of blocks in the dataset. For a finer analysis, we categorize 32 of 64 grid positions as low-density regions. Those are 50\% of regions with the least number of blocks across all layouts. 83\% of the blue pyramids appear in the low-density regions. In comparison, 88\% of all the magenta pyramids appear in regions with high block density. In~\autoref{fig:posBiasvsmAP}, we observe a slight trend implying a negative correlation [with a Pearson correlation coefficient  $r(64)\texttt{=}\texttt{-}0.3263,p\texttt{<}.008$] between the mAP and the number of block instances, which indicates high-density regions do not necessarily result in better object detection. Although the results in \autoref{fig:posBias} do not indicate a high correlation between the input distribution and the precision of the model, we do observe the lowest mAP in the (4,3) grid position due to occlusion. Since most of the block occurrences are in the (4,4) grid position, we hypothesize that certain blocks are more likely to be misclassified due to occlusion by the surrounding blocks. Our observations surrounding the increase in precision based on high-density regions contradicts the statistical analysis of the results (mAP is inversely proportional to the region's density). We, therefore, cannot deduce that one of these factors is the sole contributor to the lower mAP; however, we can attribute the low precision to a combination of the following: \textbf{1)} Blocks appearing in low-density regions could be misclassified due to a lack of many negative examples in those regions; \textbf{2)} Occlusion from surrounding blocks reduces the precision; \textbf{3)} Blocks appearing in high-density regions are also misclassified as observed from the negative correlation between the block instance counts in those regions and their resulting mAP.

In summary, the results show that Retinanet is able to localize and classify the object in our synthetic dataset with an mAP of 91.3\% for all objects. The results also show that class imbalance and bias in the spatial distribution of the training data could influence the performance of the auxiliary tasks, though the effect is not very pronounced, and there is no clear mapping from the number of instances of a class to the detection mAP for this class or the number of objects in a region and the mAP for this region.

\subsection{Learning Visuomotor Abilities with Auxiliary Tasks}
\label{sect:ablation}

We evaluate the model's primary visuomotor task of reaching for a target object in scenes with multiple objects. More specifically, we evaluate the model's output for a given visual input and natural language object description, against a given joint configuration that enables reaching for the described object.
We also evaluate the effect of training the primary visuomotor task simultaneously with the auxiliary tasks of object localization and classification. The introduction of different outputs as auxiliary tasks is hypothesized to improve the overall learning of all tasks. We evaluate the visuomotor output independently and in combination with other outputs. Our model has three output layers, one for the primary visuomotor task and two for the auxiliary tasks of object classification and localization.
The output layer producing the joint angles for the visuomotor task cannot be removed since it resolves the main objective of our task. The remaining output layers are the Retinanet classifier for identifying the objects in the image and the Retinanet regressor for localizing the objects. We experiment with all combinations of outputs resulting in a total of four combinations. We monitor the MSE of the visuomotor reach-for-grasp task when trained alone, together with classification only, with localization only and with both auxiliary tasks together. Each combination of outputs was repeated three times. As a backbone, we use the best-performing Resnet152 pre-trained on ImageNet. All tasks were trained in concert with a combined loss:
\begin{equation}
    Loss = \lambda_1\textrm{\textit{SL1}}_{reg.} + \lambda_2\textrm{\textit{FL}}_{class.} + \lambda_3\textrm{\textit{MSE}}_{vis.}
\end{equation}
Where $\lambda$ indicates the scaling factor for each loss. If a task is not included in the training, the loss is set to zero. We set all the scaling factors to 1. $SL1$ loss represents the smooth L1 regression loss for computing the localization error. The $FL$ represents the focal loss for classifying the detected objects. The MSE represents the regression loss applied to our visuomotor subnetwork for producing the robot arm configuration. We observe the MSE as a metric for the model's performance.

\begin{figure}
  \centering
  \includegraphics[width=0.96\linewidth,trim={85 140 65 80},clip]{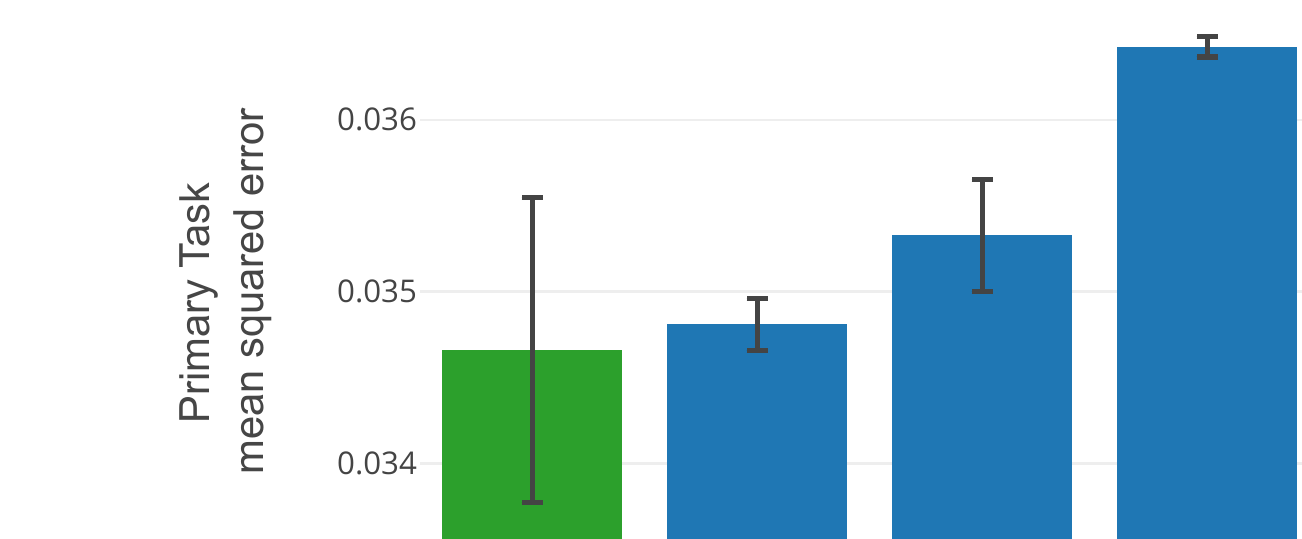}
\resizebox{\columnwidth}{!}{
\begin{tabular}{l|c|c|c|c|c}
\hline
\multicolumn{2}{l|}{\textbf{Visiomotor subnetwork (MSE)}} & \multicolumn{1}{c|}{0.0347 $\pm$ 889$\mu$} & \multicolumn{1}{c|}{0.0348 $\pm$ 150$\mu$} &
\multicolumn{1}{c|}{0.0353 $\pm$ 326$\mu$} & \multicolumn{1}{c}{0.0364 $\pm$ 60$\mu$} \\ \hline
\multicolumn{2}{l|}{\textbf{Classification subnetwork (FL)}} & \multicolumn{1}{c|}{0.3139 $\pm$ 0.003} & \multicolumn{1}{c|}{-} &
\multicolumn{1}{c|}{0.3666 $\pm$ 0.008} & \multicolumn{1}{c}{-} \\ \hline
\multicolumn{2}{l|}{\textbf{Regression subnetwork (SL1)}} & \multicolumn{1}{c|}{-} & 
\multicolumn{1}{c|}{-} &
\multicolumn{1}{c|}{0.6912 $\pm$ 0.051} & \multicolumn{1}{c}{0.5995 $\pm$ 0.014} \\ \hline
\end{tabular}
}
\caption{The MSE shows the accuracy of the primary visuomotor task when trained alone or simultaneously with different combinations of auxiliary tasks over three trials. From left to right: with classification only, with no auxiliary task, with classification and regression, and with regression only. Training with the classification auxiliary task achieves the best average results, closely followed by only training the visuomotor task. \textit{FL: Focal Loss (classification loss), SL1: Smooth L1 Loss (localization loss), MSE: Mean Squared Error Loss (visuomotor loss)}}
\label{fig:resultAblation}\vspace{\squeezefigs}
\end{figure}

Based on previous work~\cite{self}, the model was trained stochastically on single images for 32k iterations, using the Adam optimizer with $\beta_1$=0.9, $\beta_2$=0.999, $\epsilon$=$10^{-9}$, with an initial learning rate of $10^{-5}$ and scheduled to be reduced on the plateau of the primary task’s MSE. The scheduler checked the loss every 1k iterations. If the loss did not improve for two checks in a row, the learning rate was reduced.
\autoref{fig:resultAblation} shows that maintaining only the Retinanet classification subnetwork, for the classification auxiliary task, results in the lowest (best) MSE with a mean of 0.0347 and a standard deviation of 0.0008 for all three repetitions of the same setup. The variance in this condition is relatively high. The model excluding all auxiliary tasks achieves a similar second-lowest MSE with a mean of 0.0348 and a lower standard deviation of 0.0002 for all three repetitions of the same setup. In comparison to previous experiments with the robotic dataset, we observe an overall increased MSE, which can be attributed to the greater difficulty of the task of grasping a block in three-dimensional space as compared to an object that is placed flat on the table.

The results shown in \autoref{fig:resultAblation} indicate that learning all the weights in concert increases (worsens) the MSE. This is a possible outcome when dealing with a combination of different loss functions since they operate on different scales. The overall loss of the neural network is the summation of all losses combined. A loss function producing an error that is significantly larger relative to other loss functions skews the gradients improving the most erroneous objective. The L1 loss for the Retinanet regression (localization loss) appears to cause the effect.
We can also hypothesize that the Retinanet regression loss (localization loss) and the visuomotor regression loss are acting against each other by driving the optimizer’s gradients in opposing directions. This destructive interference~\cite{zhao2018modulation} could be a result of the different tasks of the Retinanet, which provides specific locations of all blocks in the image and the grasping task that focuses only on one selected block.

Training the visuomotor task together with the auxiliary classification task, however, achieves the best result, closely followed by only training the visuomotor task.  Though no significant synergistic effect can be achieved, we can show that simultaneous training of the main and an auxiliary task is possible.

\section{CONCLUSION}
We present a neural model inspired by neurocognitive models of human visual and visuomotor processing: Different visual and visuomotor tasks are realized by shared components. In humans, visual information is processed along a shared pathway that later splits into the dorsal \emph{where} and ventral \emph{what} pathways~\cite{kruger2013deep}. Both pathways then individually integrate more task-specific information like task-driven goal cues in the dorsal pathway.

We build our shared visuomotor model for object identification, localization, and grasping upon this principle of shared components. The model fuses a visual scene input with a neural encoding of a natural language object description and outputs the motor commands to reach-for-grasp a target object. This is a neurocognitively more plausible approach than existing models, where the successive processing steps of locating an object in a scene and then computing suitable motor commands are decoupled. We demonstrated the model's ability to learn the visuomotor task of reaching for a target object from a scene with multiple distractor objects through end-to-end learning in previous work. 
The advantage of such an end-to-end learning approach is that there is no need for hand-crafted, individual components, and feature extractors - all levels of the model learn to perform a given task. In this article, we present the extension of the model for a three-dimensional grasping task.

In our initial study~\cite{self}, we compared the systematic bias of the end-to-end trained model towards visually salient objects with findings from early visuomotor development in children. Despite these biases, we successfully evaluated the model on a physical robot. We attribute these biases to the limited amount of training data that could be collected with a physical robot. By using a novel, larger dataset, we can compensate for these biases, similar to a biological system accumulating experience.
However, learning complex visuomotor abilities in a physical environment from a limited set of experiences is a challenge that both biological as well as artificial systems face. Artificial systems can learn some tasks from existing datasets, e.g., image classification or object localization. Likewise, for a biological system, such purely visual tasks are learned independently and can be hypothesized to benefit the learning of more complex visuomotor skills.

We utilize this idea in the presented study and evaluate training the main visuomotor task of reaching for an object with the auxiliary vision tasks of object classification and localization. To this end, we create a fully annotated, large-scale dataset utilizing a simulation environment and augmented reality.
Sharing architectural components between different tasks is biologically plausible and efficient in terms of reducing computations and redundancy. We evaluate if it also enhances the learning for the primary tasks. A critical analysis of the results shows that some auxiliary tasks could interfere in a negative way with the primary task. This catastrophic interference is observed for training the object localization auxiliary task in the regression subnetwork alongside the primary task. However, we also observe that auxiliary training of classification in the respective subnetwork leads to a comparable or slightly better training outcome for the main visuomotor task.
This result encourages further research into shared architectures and multi-task training, especially with the aim to learn from biological systems that seem to be able to integrate different tasks without negative interference.
As an added benefit, the auxiliary tasks can serve as an analytical tool. We use the auxiliary tasks of object classification and localization, which together form the task of object detection, for analyzing whether systematic biases in detecting objects exist. The analysis is performed by calculating the mean average precision (mAP) of the object detection outputs. In contrast to our previous study~\cite{self}, we can directly observe if there is a systematic bias in visually detecting objects with certain visual properties (shape or color) or at certain positions. In our previous work, we could only observe the final visuomotor behavior of the model, with no direct indication as to whether the visual or the motor processing stream contributed to the grasp error. Being able to directly ``peak" into the outcomes of the visual stream's auxiliary tasks allowed us to analyze different error contributors. Our analysis indicated no clear biases, although the dataset has imbalances with regard to object classes and positions. We hypothesize that other factors, like occlusions due to blocks frequently stacked in regions surrounding certain locations, result in higher misclassifications. Such irregularities are not easy to remedy, since different block layout combinations are associated with natural language commands describing the blocks' properties and locations. A number of soft (e.g., equivalent number of blocks in each region) and hard (e.g., equivalent number of block class samples) constraints need to be applied to balance the dataset. Due to the variations in block layouts per sample, undersampling or oversampling techniques would not resolve this imbalance. Instead, we can reformulate the data creation task as a multivariate optimization problem. In future work, we will introduce constraints on the number of samples per region and the balance between classes to find optimal layout combinations.

In future work, we will also enhance the benefit from learning with auxiliary tasks by optimizing the weighting between different task losses and applying modulation via task-specific masks~\cite{zhao2018modulation} to avoid destructive interference between tasks. Furthermore, we will focus on the natural encoding of language descriptions of the object and evaluate this encoding via a Transformer Encoder as an additional auxiliary task.
Finally, we will use the presented model and the dataset as a stepping stone to realizing a three-dimensional reach-for-grasp task on a physical robot.
While a model trained on artificial data can not be expected to work flawlessly in a real-world scenario, there are ways to benefit from the existing dataset. We will train a neural network model trained on the large artificial dataset and then fine-tune the trained model with a much smaller set of annotated real-world images.

In summary, we present a shared visuomotor model for end-to-end learning of a complex visuomotor task.  As the main contribution, we successfully extend the model from reaching for objects on a planar surface to a three-dimensional reach for grasp task. We also evaluate the effect of biases in the training data and the effect of learning from auxiliary tasks. The latter is achieved by a multi-task and end-to-end model with shared and task-specific components, similar to models found in biological systems. We show that the primary visuomotor task and one of the two auxiliary tasks can be successfully trained simultaneously.

\bibliographystyle{IEEEtranS}
\bibliography{library,IEEEabrv}

\end{document}